# A Geometric Traversal Algorithm for Reward-Uncertain MDPs

**Eunsoo Oh**
Department of Computer Science
KAIST
Daejeon, Korea
esoh@ai.kaist.ac.kr

**Kee-Eung Kim**
Department of Computer Science
KAIST
Daejeon, Korea
kekim@cs.kaist.ac.kr

## Abstract

Markov decision processes (MDPs) are widely used in modeling decision making problems in stochastic environments. However, precise specification of the reward functions in MDPs is often very difficult. Recent approaches have focused on computing an optimal policy based on the minimax regret criterion for obtaining a robust policy under uncertainty in the reward function. One of the core tasks in computing the minimax regret policy is to obtain the set of all policies that can be optimal for some candidate reward function. In this paper, we propose an efficient algorithm that exploits the geometric properties of the reward function associated with the policies. We also present an approximate version of the method for further speed up. We experimentally demonstrate that our algorithm improves the performance by orders of magnitude.

## 1 Introduction

Markov Decision Processes (MDPs) have been a popular model for stochastic sequential decision making problems (Puterman 1994). Once we obtain the MDP model of the problem, we can solve it efficiently using various methods such as value iteration, policy iteration, or linear programming (LP) solver. However, the specification of model parameters, the transition probabilities and the rewards, can be a difficult task. For example, since the transition probabilities are estimated from the data or specified by the domain expert, there inevitably exists uncertainty regarding the accuracy of the estimation, which affects the real performance of the optimal policy obtained from the model. The specification of rewards poses a greater challenge in the sense that the current practice involves manual specification by the domain expert, and estimating it from the data still remains as an emerging research area, e.g., inverse reinforcement learning (Ng and Russell 2000).

Over the recent years, a significant body of research has dealt with finding the *robust solution* to MDPs with imprecisely specified parameters. While most of the work in the past uses the *maximin* criterion to address the uncertainty in the transition probabilities only (White and Eldeib 1994; Nilim and Ghaoui 2005) or both (White and Eldeib 1986; Givan et al. 1997), some recent approaches solely focus on the imprecise specification of rewards (Delage and Mannor 2009; Xu and Mannor 2009; Regan and Boutilier 2009; Regan and Boutilier 2010). This is not only because the rewards are more difficult to specify than the transition probabilities, but also the uncertainty in the rewards is an important subject for preference elicitation algorithms (Regan and Boutilier 2009). The *minimax regret* criterion was also proposed for computing the robust policy for MDPs with imprecise rewards (Xu and Mannor 2009; Regan and Boutilier 2009; Regan and Boutilier 2010). The challenge is in finding an effective algorithm to address the intractability result for computing the minimax regret policy.

In this paper, we propose an efficient algorithm for computing the minimax regret policy. Specifically, we build on the $\pi$Witness algorithm by Regan and Boutilier (2010) for computing the set of *nondominated policies* that are optimal for some reward function. Instead of using linear programming (LP) in finding nondominated policies, we leverage the condition on the reward function that makes a nondominated policy optimal. This technique allows the running time of our method to become linear in the number of nondominated policies. Since the number of nondominated policies is a dominant factor for the running time of the $\pi$Witness algorithm, our method is orders of magnitude faster. We also propose an anytime algorithm based on the same technique and empirically compare it with the anytime version of the $\pi$Witness algorithm.

## 2 Background

In this section, we cover the fundamentals of MDPs, reward-uncertain MDPs, and the definition of the minimax regret criterion for reward-uncertain MDPs. We also overview the previous work on computing the minimax regret policy of reward-uncertain MDPs.

### 2.1 Markov Decision Processes (MDPs)

An MDP with an infinite-horizon, finite state and action spaces is defined as $\langle S, A, T, r, \alpha, \gamma \rangle$, with the set $S$ of $n$ states, the set $A$ of $m$ actions, the state transition function $T$ which is defined as $T(s, a, s') = Pr(s'|s, a)$, the reward function $r(s, a)$, the initial state distribution $\alpha$, and the discount factor $\gamma \in [0, 1)$.

A (deterministic) policy in MDPs is a mapping $\pi : S \to A$ with the associated value function $V^\pi$ defined as

$$V^\pi(s) = r(s, \pi(s)) + \gamma \sum_{s' \in S} T(s, a, s') V^\pi(s') \quad (1)$$

which is the expected discounted sum of rewards for each state when executing the policy $\pi$. Using vector notations, the reward function is denoted as an $nm$-dimensional vector $\mathbf{r}$, and the transition function is denoted as an $nm \times n$ matrix $\mathbf{T}$. We also use $n$-dimensional vector $\mathbf{r}_\pi$ and $n \times n$ matrix $\mathbf{T}_\pi$ to denote the reward and transition functions for policy $\pi$. We can then rewrite Equation (1) as

$$\mathbf{V}^\pi = \mathbf{r}_\pi + \gamma \mathbf{T}_\pi \mathbf{V}^\pi. \quad (2)$$

The Q-function $Q : S \times A \to \mathbb{R}$ represents the expected value of taking action $a$ followed by the execution of policy $\pi$. Using vector notation, it is defined as

$$\mathbf{Q}_a^\pi = \mathbf{r}_a + \gamma \mathbf{T}_a \mathbf{V}^\pi \quad (3)$$

Each policy $\pi$ has an associated *occupancy frequency* $\mathbf{f}^\pi$ defined as

$$f^\pi(s, a) = \sum_{s'} \alpha(s') \sum_{t=0}^{\infty} \gamma^t Pr_\pi(S_t = s, A_t = a | S_0 = s')$$

which is the total discounted probability of being in state $s$ and executing action $a$. $\mathbf{f}^\pi$ can be obtained from the equation

$$\gamma \mathbf{E}_\pi^\top \mathbf{f}'^\pi + \alpha = \mathbf{0},$$

where $\mathbf{E}_\pi$ is an $n \times n$ matrix with

$$E_\pi(s, s') = \begin{cases} T(s, \pi(s), s') & \text{if } s \neq s' \\ T(s, \pi(s), s') - \frac{1}{\gamma} & \text{if } s = s' \end{cases}$$

and $\mathbf{f}'^\pi$ is an $n$-dimensional vector so that

$$f^\pi(s, a) = \begin{cases} f'^\pi(s) & \text{if } a = \pi(s) \\ 0 & \text{otherwise} \end{cases}$$

On the other hand, we can obtain $\pi$ from $\mathbf{f}_\pi$ using

$$\pi(s, a) = \frac{f^\pi(s, a)}{\sum_{a'} f^\pi(s, a')}$$

for a randomized policy, and $\pi(s) = \arg\max_a f^\pi(s, a)$ for a deterministic policy. In addition, the expected discounted sum of rewards by following policy $\pi$ can be obtained using

$$\alpha^\top \mathbf{V}^\pi = \mathbf{r}^\top \mathbf{f}_\pi. \quad (4)$$

Thus, we can interchangeably use the terms policy and occupancy frequency.

If we generalize $\mathbf{E}_\pi$ to cover the set of all policies,

$$E(sa, s') = \begin{cases} T(s, a, s') & \text{if } s \neq s' \\ T(s, a, s') - \frac{1}{\gamma} & \text{if } s = s' \end{cases}$$

then any $\mathbf{f}$ satisfying the equation

$$\gamma \mathbf{E}^\top \mathbf{f} + \alpha = \mathbf{0}$$

is called a *valid occupancy frequency* since it is associated to some policy. Hence, if we let $F$ be the set of all valid occupancy frequencies, it corresponds to the set of all possible policies.

The relationship between $\pi$ and $\mathbf{f}_\pi$ becomes more evident when we use linear programming (LP) to solve MDPs. The primal LP is formulated

$$\min_{\mathbf{V}} \quad \alpha^\top \mathbf{V}$$
$$\text{s.t.} \quad V(s) \geq r(s, a) + \gamma \sum_{s'} T(s, a, s') V(s') \ \forall s, a$$

of which the dual LP is formulated as

$$\max_{\mathbf{f}} \quad \mathbf{r}^\top \mathbf{f}$$
$$\text{s.t.} \quad \gamma \mathbf{E}^\top \mathbf{f} + \alpha = \mathbf{0} \quad (5)$$
$$\mathbf{f} \geq \mathbf{0}.$$

Note that the solution of the dual LP is the occupancy frequency of the optimal policy.

### 2.2 Reward-Uncertain MDPs and Minimax Regret Policies

The exact specification of the reward function can be difficult in practice. The reward-uncertain MDP (RUMDP) (Regan and Boutilier 2010) extends the standard MDP by allowing a set of feasible reward functions instead of requiring a single exact reward function. The RUMDP is defined by $\langle S, A, T, R, \alpha, \gamma \rangle$ where $R$ is the space of feasible reward functions, replacing the reward function $r$ in the standard MDP. Hence, the uncertainty in the reward function is confined to the space $R$. In addition, following the original work on RUMDPs, we assume that $R$ is a bounded and convex polytope defined by the set of linear constraints $\{\mathbf{r} | \mathbf{Ar} \leq \mathbf{b}\}$. We use $|R|$ to denote the number of constraints (the number of rows in $\mathbf{A}$) and $\dim(R)$ to denote the dimension of $R$.[1]

Given an RUMDP, we want to find a policy that is *robust* to the uncertainty in the reward function. We thus require a criterion to compare the robustness of policies. We adopt the *minimax regret criterion* (Boutilier et. al. 2006, Xu and Mannor 2009, Regan and Boutilier 2009, Regan and Boutilier 2010), which is defined as

$$MMR(R) = \min_{\mathbf{f} \in F} \max_{\mathbf{g} \in F} \max_{\mathbf{r} \in R} \mathbf{r}^\top \mathbf{g} - \mathbf{r}^\top \mathbf{f}. \quad (6)$$

The formulation can be viewed as if there is an adversary choosing the reward function $\mathbf{r}$ to maximize the loss with respect to the optimal policy (i.e. occupancy frequency) $\mathbf{g}$, while our agent is selecting policy $\mathbf{f}$ to minimize the loss. The minimax regret corresponds to the worst-case bound on the loss. Formally, let $\mathbf{f}_{MMR(R)} = \arg\min_{\mathbf{f}} \max_{\mathbf{g}} \max_{\mathbf{r}} \mathbf{r}^\top \mathbf{g} - \mathbf{r}^\top \mathbf{f}$ be the minimax regret policy and $MMR(R)$ be the minimax regret achieved. Then, given any realization of $\mathbf{r}$, no policy can outperform $\mathbf{f}$ by more than $MMR(R)$ in terms of the expected value.

### 2.3 Computing Minimax Regret Policies

Although the minimax regret is a natural criterion for robustness, computing the minimax policy for a RUMDP is known to be NP-hard (Xu and Mannor 2009). A number of approaches exist, but we only review those involving *nondominated policies*.

A policy $\mathbf{g}$ is defined to be *nondominated* with respect to the feasible reward space $R$ if and only if

$$\exists \mathbf{r} \in R \text{ s.t. } \mathbf{r}^\top \mathbf{g} \geq \mathbf{r}^\top \mathbf{f}, \quad \forall \mathbf{f} \in F,$$

which states that a nondominated policy should be optimal for some feasible reward function. Let $\Gamma$ denote the set of nondominated policies. The set $\Gamma$ is useful for computing the minimax policy in equation (6) because, for any $\mathbf{f} \in F$, $(\arg\max_{\mathbf{g} \in F} \max_{\mathbf{r} \in R} \mathbf{r}^\top \mathbf{g} -$ $\mathbf{r}^\top \mathbf{f}) \in \Gamma$. We can verify this property from the fact that the policy $\mathbf{g}$ is chosen to maximize the expected value for some reward $\mathbf{r} \in R$. In addition, if we view the minimax regret policy as a randomization over a set of deterministic policies, the probability of choosing a dominated policy can be made be zero since there should be a nondominated policy that performs better.

---
[1] Generally speaking, $\dim(R)$ is equal to the dimension of $\mathbf{r}$, i.e., $|S| \times |A|$. In some cases, however, the reward function can be compactly represented in a factored form, so that $\dim(R)$ is proportional to the number of basis functions and significantly less than $|S| \times |A|$.

Xu and Mannor (2009) propose the following LP formulation for computing the minimax regret policy of RUMDP with $R = \{\mathbf{r} | \mathbf{Ar} \leq \mathbf{b}\}$ using the set $\Gamma$ of nondominated policies:

$$\begin{aligned}
\min_{\mathbf{z}, \mathbf{c}, \delta} \quad & \delta \\
\text{subject to} \quad & \sum_{i=1}^{|\Gamma|} c_i = 1 \\
& \mathbf{c} \geq \mathbf{0} \\
& \delta \geq \mathbf{b}^\top \mathbf{z}_i \\
& \mathbf{A}^\top \mathbf{z}_i + \hat{\Gamma} \mathbf{c} = \mathbf{g}_i \\
& \mathbf{z}_i \geq \mathbf{0}
\end{aligned} \bigg\} i = 1, \ldots, |\Gamma|$$

where $\hat{\Gamma}$ is the matrix formed by taking the elements in $\Gamma$ as columns. The $|\Gamma|$-dimensional vector $\mathbf{c}$ represents the convex combination of nondominated policies in the minimax regret policy. The last three constraints are from the dual of

$$\begin{aligned}
\max_{\mathbf{r}} \quad & \mathbf{r}^\top (\mathbf{g}_i - \mathbf{c}^\top \hat{\Gamma}) \\
\text{subject to} \quad & \mathbf{Ar} \leq \mathbf{b}
\end{aligned}$$

for each adversarial policy $\mathbf{g}_i \in \Gamma$. The overall LP has $O(|R||\Gamma|)$ variables and $O(|\Gamma|)$ constraints.

Since $|\Gamma|$ can be very large, Regan and Boutilier (2010) propose ICG-ND, a technique based on constraint generation for LPs. Given the subset GEN of possible adversarial policies and rewards, the minimax regret policy can be found by LP

$$\begin{aligned}
\min_{\mathbf{f}, \delta} \quad & \delta \\
\text{subject to} \quad & \mathbf{r}_i^\top \mathbf{g}_i - \mathbf{r}_i^\top \mathbf{f} \leq \delta \quad \forall \langle \mathbf{g}_i, \mathbf{r}_i \rangle \in \text{GEN} \\
& \gamma \mathbf{E}^\top \mathbf{f} + \alpha = \mathbf{0}
\end{aligned}$$

In order to iteratively generate the set GEN, an LP is solved for each $\mathbf{g} \in \Gamma$ to determine the reward that maximizes the regret for the current solution $\mathbf{f}$:

$$\begin{aligned}
\min_{\mathbf{r}} \quad & \mathbf{r}^\top \mathbf{g} - \mathbf{r}^\top \mathbf{f} \\
\text{subject to} \quad & \mathbf{Ar} \leq \mathbf{b}
\end{aligned}$$

The $\mathbf{g}$ with the largest objective value is regarded as the maximally violated constraint, and included into GEN along with the associated $\mathbf{r}$.

While it is evident from ICG-ND that the set $\Gamma$ of nondominated policies is important in computing minimax regret policies, enumerating the set $\Gamma$ still remains as a challenge. Hence, Regan and Boutilier (2010) propose the $\pi$Witness algorithm for the exact computation of $\Gamma$. The algorithm incrementally constructs the

**Algorithm 1:** The $\pi$Witness Algorithm

**begin**
$\quad$ $\mathbf{r} \leftarrow$ some arbitrary $\mathbf{r} \in R$
$\quad$ $\mathbf{f} \leftarrow \textbf{findOptPolicy}(\mathbf{r})$
$\quad$ $\Gamma \leftarrow \{\mathbf{f}\}$
$\quad$ agenda $\leftarrow \{\langle \mathbf{r}, \mathbf{f} \rangle\}$
$\quad$ **while** *agenda is not empty* **do**
$\quad\quad$ $\mathbf{f} \leftarrow$ next item in agenda
$\quad\quad$ **for** $s, a$ **do**
$\quad\quad\quad$ $\mathbf{r}' \leftarrow \textbf{findWitnessRewardFn}(\mathbf{f}, s, a, \Gamma)$
$\quad\quad\quad$ **while** *witness found* **do**
$\quad\quad\quad\quad$ $\mathbf{f}' \leftarrow \textbf{findOptPolicy}(\mathbf{r}')$
$\quad\quad\quad\quad$ add $\mathbf{f}'$ to $\Gamma$
$\quad\quad\quad\quad$ add $\langle \mathbf{r}', \mathbf{f}' \rangle$ to agenda
$\quad\quad\quad\quad$ $\mathbf{r}' \leftarrow \textbf{findWitnessRewardFn}(\mathbf{f}, s, a, \Gamma)$

set of nondominated policies by finding a reward function that makes some policy optimal but not yet in $\Gamma$. This method is analogous to the witness algorithm in the POMDP literature. Algorithm 1 presents the pseudo-code of the algorithm.

**findOptPolicy**($\mathbf{r}$) computes an optimal policy for the MDP with reward $\mathbf{r}$. We can use any MDP algorithm for this procedure, e.g., value iteration, policy iteration, or LP.

**findWitnessRewardFn**($\mathbf{f}, s, a, \Gamma$) attempts to find $\mathbf{r}$ for which a local adjustment of $\mathbf{f}$ at $(s, a)$ has higher value than any $\mathbf{f}' \in \Gamma$ by solving an LP. This LP has $O(|\Gamma|)$ constraints and $O(|S||A|)$ variables.

Its running time is polynomial in $|\Gamma|$, $|S|$, and $|A|$, but not linear in $|\Gamma|$. Since $|\Gamma| \gg |S||A|$ in general, the size of $\Gamma$ is a dominant factor in running time of the $\pi$Witness algorithm. In the later section, we empirically show that computing $\Gamma$ takes much more time than ICG-ND alone. Hence, improvement in computing $\Gamma$ is critical to the efficient computation of the minimax regret policies in RUMDPs.

## 3 Geometric Traversal for Nondominated Policies

In this section, we present our algorithm for computing the set of nondominated policies in RUMDPs, leveraging the optimality condition of reward function in MDPs.

### 3.1 Optimality Condition for Rewards

Consider an MDP $M = \langle S, A, T, \mathbf{r}^*, \alpha, \gamma \rangle$ with an optimal policy $\pi$. If the change $\Delta \mathbf{r}$ in the reward function $\mathbf{r}^*$ is very small, $\pi$ may still remain as an optimal policy. Treating the reward function $\mathbf{r} = \mathbf{r}^* + \Delta \mathbf{r}$ as a variable vector, we can obtain a necessary and sufficient condition for $\mathbf{r}$ that guarantees the optimality of $\pi$ using the result in (Ng and Russell 2000):

$$\mathbf{V}^\pi \geq \mathbf{Q}_a^\pi \quad \forall a.$$

Using equations (2) and (3), the above inequality becomes

$$\mathbf{r}_\pi + \gamma \mathbf{T}_\pi \mathbf{V}^\pi \geq \mathbf{r}_a + \gamma \mathbf{T}_a \mathbf{V}^\pi \quad \forall a. \quad (7)$$

Note also that, from equation (2),

$$\mathbf{V}^\pi = (\mathbf{I} - \gamma \mathbf{T}_\pi)^{-1} \mathbf{r}_\pi = (-\gamma \mathbf{E}_\pi)^{-1} \mathbf{r}_\pi.$$

Hence equation (7) becomes $\forall a$,

$$\mathbf{r}_\pi - \gamma \mathbf{T}_\pi (\gamma \mathbf{E}_\pi)^{-1} \mathbf{r}_\pi \geq \mathbf{r}_a - \gamma \mathbf{T}_a (\gamma \mathbf{E}_\pi)^{-1} \mathbf{r}_\pi$$
$$\Leftrightarrow \mathbf{r}_\pi + (\mathbf{I} - \gamma \mathbf{T}_\pi)(\gamma \mathbf{E}_\pi)^{-1} \mathbf{r}_\pi$$
$$\geq \mathbf{r}_a + (\mathbf{I} - \gamma \mathbf{T}_a)(\gamma \mathbf{E}_\pi)^{-1} \mathbf{r}_\pi$$
$$\Leftrightarrow \mathbf{r}_\pi - \gamma \mathbf{E}_\pi (\gamma \mathbf{E}_\pi)^{-1} \mathbf{r}_\pi \geq \mathbf{r}_a - \gamma \mathbf{E}_a (\gamma \mathbf{E}_\pi)^{-1} \mathbf{r}_\pi.$$

Since the left-hand side $\mathbf{r}_\pi - \gamma \mathbf{E}_\pi (\gamma \mathbf{E}_\pi)^{-1} \mathbf{r}_\pi = \mathbf{0}$, we obtain

$$\mathbf{r} - \mathbf{E}(\gamma \mathbf{E}_\pi)^{-1} \mathbf{r}_\pi \leq \mathbf{0}.$$

Using the definitions $\mathbf{r} = \mathbf{r}^* + \Delta \mathbf{r}$ and $\mathbf{r}_\pi = \mathbf{r}_\pi^* + \Delta \mathbf{r}_\pi$, we have

$$\mathbf{r}^* + \Delta \mathbf{r} - \mathbf{E}(\gamma \mathbf{E}_\pi)^{-1}(\mathbf{r}_\pi^* + \Delta \mathbf{r}_\pi) \leq \mathbf{0}. \quad (8)$$

We refer to equation (8) as the *reward optimality condition* with respect to policy $\pi$. It yields the set of linear inequalities for the changes in the reward function $\Delta r(s, a)$ that defines the boundary around $\mathbf{r}^*$ where $\pi$ remains as optimal.

The reward optimality condition is essentially equivalent to performing sensitivity analysis on LP (Bertsimas and Tsitsiklis 1997) for solving the MDP in equation (5). However, we cannot leverage LP solvers for computing the exact boundary since their sensitivity analysis results project the system of inequalities into each state and action, i.e., the change in the reward function is analyzed independently for each state and action.

### 3.2 Geometric Traversal Algorithm

The reward optimality condition in equation (8) is essentially a set of inequalities, each of which describes a hyperplanar boundary. Note that the region defined by the reward optimality condition is a convex bounded polytope. The space of feasible reward functions is visualized in Figure 1 (a). The space is partitioned into the regions defined by the reward optimality condition, each region corresponding to a nondominated policy. Hence, we can view the partitioned space as a connected undirected graph with nodes corresponding to nondominated policies and edges corresponding to the adjacency of their optimal reward regions, as shown in Figure 1 (b).

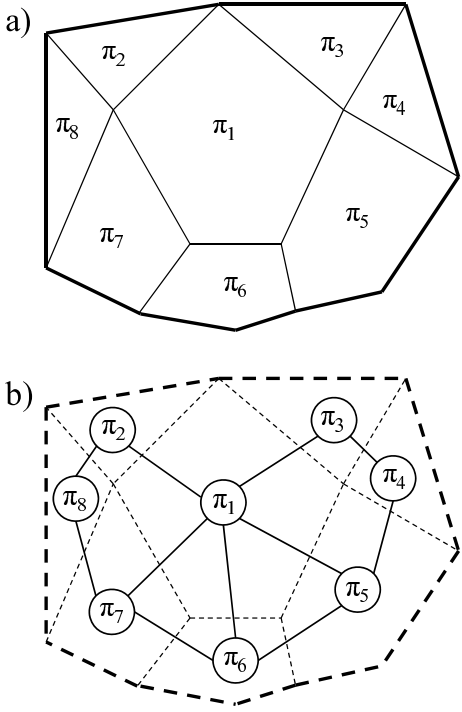

Figure 1: An example of 2-dimensional feasible reward function space $R$ (bold polygon) and its partition induced by nondominated policies. (a) Reward regions (polygons) of each nondominated policy in $R$. (b) Graph representation of the nondominated policies.

Our geometric traversal algorithm for finding nondominated policies essentially constructs this graph using the reward optimality condition. Since all the nodes are connected, any exhaustive traversal algorithm can be used, e.g., breadth-first or depth-first. Algorithm 2 presents the pseudo-code of the algorithm.

**findRewardOptRgn**$(\mathbf{r}, \mathbf{f})$ constructs the set of hyperplanes defined by the reward optimality condition in equation (8). Specifically, it computes the set $H$ of hyperplanes, where each hyperplane $h \in H$ is represented as $\mathbf{c}_h^\top \mathbf{r} \leq d_h$. Hence each hyperplane can be represented as a pair $\langle \mathbf{c}_h, d_h \rangle$. Each hyperplane corresponds to one of the edges in the graph of nondominated policies.

**findAdjRewardFn**$(h, H)$ yields a reward function which is located in the adjacent reward region across the hyperplane $h$. It is obtained by solving the following LP with a small positive constant $\delta$ for excluding

**Algorithm 2:** Geometric Traversal Algorithm
**begin**
$\quad \mathbf{r} \leftarrow$ some arbitrary $\mathbf{r} \in R$
$\quad \mathbf{f} \leftarrow$ **findOptPolicy**$(\mathbf{r})$
$\quad \Gamma \leftarrow \{\mathbf{f}\}$
$\quad$ agenda $\leftarrow \{\langle \mathbf{r}, \mathbf{f} \rangle\}$
$\quad$ **while** *agenda is not empty* **do**
$\quad\quad \langle \mathbf{r}, \mathbf{f} \rangle \leftarrow$ next item in agenda
$\quad\quad H \leftarrow$ **findRewardOptRgn**$(\mathbf{r}, \mathbf{f})$
$\quad\quad$ **for** $h \in H$ **do**
$\quad\quad\quad \mathbf{r}' \leftarrow$ **findAdjRewardFn**$(h, H)$
$\quad\quad\quad$ **if** $\mathbf{r}'$ *is found* **then**
$\quad\quad\quad\quad \mathbf{f}' \leftarrow$ **findOptPolicy**$(\mathbf{r}')$
$\quad\quad\quad\quad$ **if** $\mathbf{f}' \notin \Gamma$ **then**
$\quad\quad\quad\quad\quad$ add $\mathbf{f}'$ to $\Gamma$
$\quad\quad\quad\quad\quad$ add $\langle \mathbf{r}', \mathbf{f}' \rangle$ to agenda

points exactly on $h$:

$$\max_{\mathbf{r}'} \quad 0$$
$$\text{s.t.} \quad \mathbf{A}\mathbf{r}' \leq \mathbf{b}$$
$$\left.\begin{array}{l} \mathbf{c}_{h'}^\top \mathbf{r}' \leq d_{h'} \text{ if } h' \neq h \\ \mathbf{c}_{h'}^\top \mathbf{r}' \geq d_{h'} + \delta \text{ if } h' = h \end{array}\right\} \forall h' \in H \quad (9)$$

Note that the reward region of interest is adjacent to the current reward region by $h$, since we reverse the direction of the inequality for $h$ while keeping every other $h' \in H$ unchanged. If the above LP yields a feasible solution, it implies that there exists an adjacent reward region with potentially a new nondominated policy.

Upon the termination of the geometric traversal algorithm, $\Gamma$ will be the complete set of nondominated policies. If we implement $\Gamma$ using a hash set, all the set operations used in the algorithm take $O(1)$ time.

The running time of our geometric traversal algorithm is polynomial in $|S|$ and $|A|$, and *linear* in $|\Gamma|$ since:

- Each call to **findRewardOptRgn**$(\mathbf{r}, \mathbf{f})$ takes $O(|S|^2|A|\dim(R))$ and is called $|\Gamma|$ times. Note that the running time of the procedure is independent of $|\Gamma|$.

- Since the size of each $H$ is at most $|S||A|$, **findAdjRewardFn**$(h, H)$ is called at most $|S||A|$ times for each $\mathbf{f} \in \Gamma$, hence it is called a total of $|\Gamma||S||A|$ times. Each call to the procedure requires solving an LP with $\dim(R)$ variables and $|H| \leq |S||A|$ constraints, of which the running time is independent of $|\Gamma|$.

- **findOptPolicy**$(\mathbf{r})$ is called only when an adjacent reward function is found, so it is called

**Algorithm 3:** Approximate Geometric Traversal Algorithm

**begin**
  $\Gamma \leftarrow \{\}$
  agenda $\leftarrow \{\}$
  **while** $\Gamma$ *is not sufficiently gathered* **do**
    $\mathbf{r} \leftarrow$ some arbitrary $\mathbf{r} \in R$
    $l \leftarrow$ arbitrary straight line passing through $\mathbf{r}$
    $\mathbf{f} \leftarrow$ **findOptPolicy**$(\mathbf{r})$
    add $\mathbf{f}$ to $\Gamma$
    add $\langle \mathbf{r}, \mathbf{f} \rangle$ to agenda
    **while** *agenda is not empty* **do**
      $\langle \mathbf{r}, \mathbf{f} \rangle \leftarrow$ next item in agenda
      $H \leftarrow$ **findRewardOptRgn**$(\mathbf{r}, \mathbf{f})$
      $\{\mathbf{r}_1, \mathbf{r}_2\} \leftarrow$ find two intersections from $H$
      **for** $\mathbf{r}' \in \{\mathbf{r}_1, \mathbf{r}_2\}$ **do**
        $\mathbf{f}' \leftarrow$ **findOptPolicy**$(\mathbf{r}')$
        add $\langle \mathbf{r}', \mathbf{f}' \rangle$ to agenda
        **if** $\mathbf{f}' \notin \Gamma$ **then**
          add $\mathbf{f}'$ to $\Gamma$

$O(|\Gamma||S||A|)$ times. The running time of the procedure is again independent of $|\Gamma|$.

In short, each iteration in the while loop takes the running time polynomial in $|S|$ and $|A|$, but independent of $|\Gamma|$, so the overall time complexity of the geometric traversal algorithm is linear in $|\Gamma|$.

### 3.3 Approximate Method For Computing Nondominated Policies

Although the geometric traversal algorithm significantly improves the running time, it still can take a large amount of time since the algorithm collects every nondominated policy, potentially as many as $|A|^{|S|}$. Regan and Boutilier (2010) propose a method for computing a subset of nondominated policies, using the $\pi$Witness algorithm in an anytime manner. Using a subset of nondominated policies, they use ICG-ND to compute an approximate minimax regret policy.

Since our algorithm also incrementally constructs $\Gamma$, it can be also used in an anytime fashion to compute a subset of nondominated policies. The idea is to traverse in each iteration a subset of adjacent reward regions that are encountered while moving along a straight line. Specifically, our approximate algorithm starts with an arbitrary reward function $\mathbf{r}$ in $R$ and a random straight line $l$ that passes through $\mathbf{r}$. All the points $\mathbf{r}'$ on the line $l$ with direction vector $\mathbf{w}$ are represented by the equation $\mathbf{r}' = \mathbf{r} + \mathbf{w} \cdot t$.

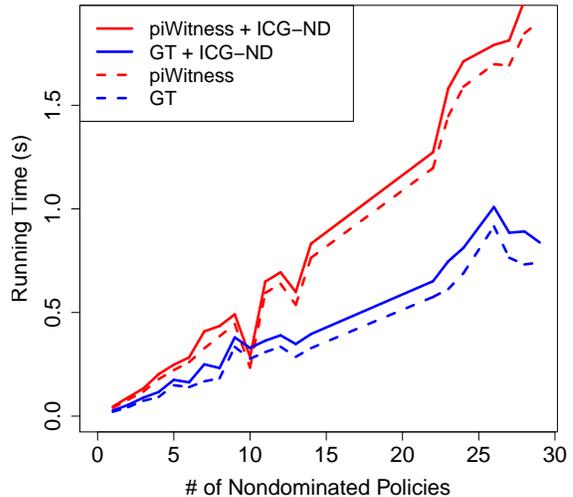

Figure 2: Running times of algorithms on 100 random instances of RUMDP with $|S| = 8, |A| = 5, \dim(R) = 2$.

Then, once we compute the set $H$ of hyperplanes defining the boundary of the current optimal reward region using **findRewardOptRgn**$(\mathbf{r}, \mathbf{f})$, we can obtain the intersection of line $l$ and hyperplane $h \in H$ by solving the system of linear equations:

$$\mathbf{r}' = \mathbf{r} + \mathbf{w} \cdot t$$
$$\mathbf{c}_h^\top \mathbf{r}' = d_h$$

Two intersections with line $l$ and the boundary defined by $H$ is obtained by taking $\mathbf{r}'$ with the minimum among the positive solutions and the maximum among the negative solutions of $t$. By adding and subtracting a small positive constant $\delta$ to the solutions, we obtain the rewards in the two adjacent reward regions. Once we gather all the adjacent reward regions along the current line $l$, we restart with an arbitrary reward function $\mathbf{r}$ and a new random straight line $l$. Algorithm 3 presents the pseudo-code of our approximate method based on the geometric traversal algorithm.

## 4 Experiments

We tested the performance of our algorithm on randomly generated instances of RUMDPs with different state sizes and reward function dimensions. For each setting of the state size and the reward function dimension, we randomly generated 100 instances of RUMDPs, following the same experimental evaluation setup in (Regan and Boutilier 2010). We ran geometric traversal (GT) algorithm, $\pi$Witness, and ICG-ND. Note that GT and $\pi$Witness are used to precom-pute the set $\Gamma$ of nondominated policies, while ICG-ND computes the minimax regret policy using $\Gamma$.

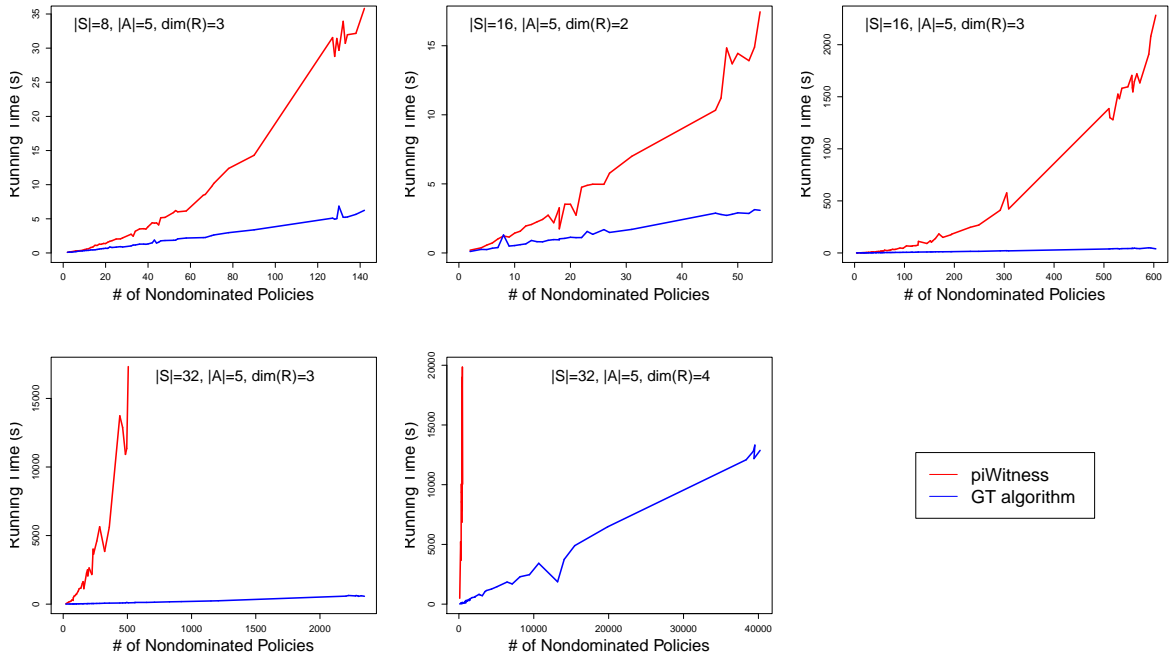

Figure 3: The running times of GT and $\pi$Witness on larger RUMDPs.

Figure 2 compares the running times of the algorithms with respect to different sizes of $\Gamma$. The figure shows the running time plots of $\pi$Witness only, GT only, $\pi$Witness with ICG-ND, and GT with ICG-ND. First, notice from the graph that whether we use $\pi$Witness or GT, computing $\Gamma$ is the dominant factor in the overall running time for computing the minimax regret policy. Second, GT is significantly faster than $\pi$Witness due to its time complexity linear in $|\Gamma|$.

When the running times are compared on larger RUMDPs, the performance gain achieved by GT is much clearer. Figure 3 shows the comparisons of running times on 5 different settings in the sizes of RUMDPs. It is evident from the graph that we typically obtain several magnitudes of order improvement in the running time using GT.

We also experimented with the approximate version of GT. Table 1 summarizes the results on RUMDPs with different sizes. For each of the 4 size settings, we generated 5 to 10 random instances of RUMDPs, and ran approximate GT and anytime $\pi$Witness algorithms until the relative error in the minimax regret falls below 10%, 5% and 1%, or time out at 20 minutes. We measured the execution time and the number of produced nondominated policies upon termination. As expected, our approximate GT signifi- cantly outperformed anytime $\pi$Witness in terms of execution time, while generating significantly more nondominated policies due to the lack of a prioritization heuristic. Developing a good heuristic for approximate GT remains as a future work.

All the algorithms were implemented in Java, and CPLEX 12.1 was used as the LP solver.

## 5 Conclusion

We have presented methods to significantly improve the speed of computing the minimax regret policies in RUMDPs. Specifically, we identified that the bottleneck of the state-of-the-art RUMDP algorithm, $\pi$Witness, is in computing the set of nondominated policies, and proposed an efficient algorithm that exploits the geometric properties of reward functions associated with nondominated policies. The end result is a linear time algorithm with respect to the number of nondominated policies in the model, achieving orders of magnitude performance improvement. We also presented an approximate version of the method which does not depend on solving any LP. Experimental results show that the approximate method outperforms anytime version of the $\pi$Witness algorithm in terms of execution time.

There are a number of future research directions worth pursuing. First, it would be useful to extend the al-gorithm to factored domains. Our approach currently requires construction of $|S||A|$ hyperplanes that define the reward optimality condition, but we conjecture that we can obtain a compact representation. Second, although we achieved significant running time improvement in anytime performance, but we currently lack a heuristic to prioritize nondominated policies, producing a significantly larger set of nondominated policies than $\pi$Witness to achieve the same level of minimax regret error. It would be interesting to investigate whether we can adapt the idea behind the Regan and Boutilier (2009) heuristic into our algorithm.

| $|S|$ | $|A|$ | $\dim(R)$ | $|\Gamma|$ | MMR error | Approximate GT | | Anytime $\pi$Witness | |
|---|---|---|---|---|---|---|---|---|
| | | | | | time (sec) | $|\hat{\Gamma}|$ | time (sec) | $|\hat{\Gamma}|$ |
| 16 | 5 | 2 | 30.5 | $< 1\%$ | 0.3 | 14.0 | 2.6 | 14.4 |
| | | | | $< 5\%$ | 0.3 | 12.7 | 2.5 | 14.3 |
| | | | | $< 10\%$ | 0.3 | 11.7 | 0.4 | 10.3 |
| 32 | 5 | 3 | 716.4 | $< 1\%$ | 4.0 | 164.2 | n/a | n/a |
| | | | | $< 5\%$ | 3.25 | 154.6 | 108.9 | 71.1 |
| | | | | $< 10\%$ | 2.82 | 137.9 | 81.0 | 61.6 |
| 64 | 5 | 3 | 1252.2 | $< 1\%$ | 26.2 | 456.8 | n/a | n/a |
| | | | | $< 5\%$ | 22.0 | 442 | 356.8 | 50.3 |
| | | | | $< 10\%$ | 14.0 | 344 | 205.5 | 44.7 |
| 64 | 5 | 4 | 10696.8 | $< 1\%$ | 32.5 | 841.6 | n/a | n/a |
| | | | | $< 5\%$ | 26.1 | 729.6 | 243.1 | 72.2 |
| | | | | $< 10\%$ | 19.2 | 575.2 | 146.8 | 55.8 |

Table 1: Experimental results of approximate GT and anytime $\pi$Witness.

**Acknowledgements**

This work was supported by the National Research Foundation of Korea (Grant# 2009-0069702) and by the Defense Acquisition Program Administration and the Agency for Defense Development of Korea (Contract# UD080042AD).

**References**

D. Bertsimas and J. Tsitsiklis (1997). *Introduction to Linear Optimization*, Athena Scientific, Belmont, Massachusetts.

C. Boutilier, R. Patrascu, P. Poupart and D. Schuumans (2006). Constraint-based Optimization and Utility Elicitation Using the Minimax Decision Criterion, *Artificial Intelligence* **170**:686-713.

E. Delage and S. Mannor (2009). Percentile Optimization for Markov Decision Processes with Parameter Uncertainty. *Operations Research* **58**:203-213.

R. Givan, S. Leach, and T. Dean (1997). Bounded Parameter Markov Decision Processes. In *Proceedings of the 4th European Conference on Planning*, 234-246.

A. Ng and S. Russell (2000). Algorithms for Inverse Reinforcement Learning. In *Proceedings of the International Conference on Machine Learning* (ICML-00), 663-670.

A. Nilim and L.E. Ghaoui (2005). Robust Control of Markov Decision Processes with Uncertain Transition Matrices. *Operations Research* **53**(5):780-798.

M. Puterman (1994). *Markov Decision Processes: Discrete Stochastic Dynamic Programming*, Wiley, Newyork.

K. Regan and C. Boutilier (2009). Regret-based Reward Elicitation for Markov Decision Processes. In *Proceedings of the 25th Conference on Uncertainty in Artificial Intelligence* (UAI-09).

K. Regan and C. Boutilier (2010). Robust Policy Computation in Reward-uncertain MDPs using Nondominated Policies. In *Proceedings of the 25th National Conference on Artificial Intelligence* (AAAI-10), Atlanta, USA.

C. C. White and H. K. Eldeib (1986). Parameter Imprecision in Finite State, Finite Action Dynamic Programs. *Operations Research* **34**:120-129.

C. C. White and H. K. Eldeib (1994). Markov Decision Processes with Imprecise Transition Probabilities. *Operations Research* **43**:739-749.

H. Xu and S. Mannor (2009). Parametric Regret in Uncertain Markov Decision Processes. In *Proceedings of the 48th IEEE Conference on Decision and Control* (CDC-09), 3606-3613.